\documentclass{article}

     \PassOptionsToPackage{numbers, compress}{natbib}

\usepackage[preprint]{neurips_2024}

\usepackage[utf8]{inputenc} 
\usepackage[T1]{fontenc}    
\usepackage{hyperref}       
\usepackage{url}            
\usepackage{booktabs}       
\usepackage{amsfonts}       
\usepackage{nicefrac}       
\usepackage{microtype}      
\usepackage{xcolor}         
\usepackage{amsmath, amssymb, amsfonts}        
\usepackage{subcaption}
\usepackage{graphicx}
\usepackage{algorithm}
\usepackage{algorithmicx}
\usepackage{algpseudocode}
\usepackage{multirow}
\usepackage{multicol}
\usepackage{array}

\newtheorem{lemma}{Lemma}

\title{Inducing Robustness in a 2-dimensional Direct Preference Optimisation Paradigm}

\author{Sarvesh Shashidhar$^{1}$, Ritik$^{1}$, Nachiketa Patil$^{1}$, Suraj Racha$^{2}$, Ganesh Ramakrishnan$^{3}$ \\
$^{1}$Centre for Machine Intelligence and Data Science, IIT Bombay \\
$^{2}$Koita Centre for Digital Health, IIT Bombay \\
$^{3}$Department of Computer Science and Engineering, IIT Bombay \\
\texttt{\{sarvesh.s, ritik, npatil\}@minds.iitb.ac.in} \\
\texttt{23d1627@iitb.ac.in , ganesh@cse.iitb.ac.in}
}

\begin{document}
\maketitle

\begin{abstract}
  \textbf{Direct Preference Optimisation (DPO)} has emerged as a powerful method for aligning Large Language Models (LLMs) with human preferences, offering a stable and efficient alternative to approaches that use Reinforcement learning via Human Feedback. In this work, we investigate the performance of DPO using open-source preference datasets. One of the major drawbacks of DPO is that it doesn't induce granular scoring and treats all the segments of the responses with equal propensity. However, this is not practically true for human preferences since even ``good'' responses have segments that may not be preferred by the annotator. To resolve this, a \textbf{2-dimensional} scoring for DPO alignment called \textbf{2D-DPO} was proposed. We explore the 2D-DPO alignment paradigm and the advantages it provides over the standard DPO by comparing their win rates. It is observed that these methods, even though effective, are not robust to label/score noise. To counter this, we propose an approach of incorporating \textbf{segment-level} score noise robustness to the 2D-DPO algorithm. Along with theoretical backing, we also provide empirical verification in favour of the algorithm and introduce other noise models that can be present.
\end{abstract}

\section{Introduction}
With the advent of Large Language Models (LLMs) and their increase in popularity, discussions and debates about the safety and correctness of model responses have become crucial. Having an LLM that gives the correct answer is not sufficient anymore - we need models that give ``safe'' and ``responsible'' answers. This idea of \textbf{aligning} LLMs to human preferences has taken priority in the community ( Liu et. al. \cite{align1},  Shen et. al. \cite{align2},  Wang et. al. \cite{align3}) and multiple methods have been proposed for the same. \\

Since most LLM responses need to be aligned to human preferences, the intuitive way to perform alignment is to include \textbf{Human Feedback} in the training process. Often termed as \textit{Reinforcement Learning with Human Feedback (RLHF)} (as proposed by Christiano et. al. \cite{rlhf}), the algorithm performs alignment usign \textbf{pairwise preferences} provided by the human annotators between the responses. The algorithm follows a two-step process, which first learns a reward model to maximise reward scores between the two responses and then models an LLM policy around that reward model. \\

Although effective, this method is computationally heavy and unstable due to the two-step optimization process. A more elegant solution was proposed by Rafailov et. al. \cite{dpo} in the form of \textbf{Direct Preference Optimisation (DPO)}. DPO performs a single-step optimisation where the LLM policy is modelled directly using the annotated preference dataset. This simplified the overall process, was computationally lighter, while being on par with RLHF when it came to performance. \\

Over the years, many extensions to DPO have been proposed, each optimising some aspect of the original algorithm. In their recent work, Li et. al. \cite{2ddpo} claimed that because DPO assigns a single score to preferred and rejected responses, it fails to capture the granularity of human responses. To correct for this drawback, they proposed \textbf{2-dimensional DPO (2D-DPO)} that divides the response into \textit{segments} and each segment is scored across \textit{5 aspects} - ``Completeness'', ``Clarity'', ``Correctness'', ``Safety'' and ``Helpfulness''. Backing their claim with empirical evidence, they showed that this paradigm improved upon the performance of Vanilla DPO.  \\

The algorithms discussed thus far are effective for preference alignment, but work under the assumption that the reward scores provided by the human annotators are the ground truth. In practical settings however, there is a significant chance that some of these labels and scores are \textit{noisy} and do not reflect the true preference of the user. Robustness to such noisy labels in vanilla DPO was explored by Chowdhury et. al. \cite{robdpo} where they handled the probabilistic flipping of labels between rejected and preferred responses. However, we were unable to find a similar paradigm for 2D-DPO. In this work, we propose a \textit{Robust 2D-DPO} framework that is robust to perturbations at the segment-level scores.  We provide theoretical formulations and empirical verifications for a framework that is robust to small, segment-level perturbations drawn uniformly from a distribution.

\section{Related Work}
\textbf{Reinforcement Learning from Human Feedback.} Emperical evidence has suggested the efficacy of incorporating human feedback and preferences in domains like robotics (Abramson et. al. \cite{robot}) and LLM training (Ziegler et. al. \cite{rlhf2}) to mention a few. In the case of LLM training, \textbf{Re-inforcement Learning using Human Feedback (RLHF)} as proposed by Christiano et. al. \cite{rlhf} has been a pivotal algorithm for preference optimisation. Using a two-step process of reward model formulation and policy modelling, RLHF has proven effective in optimising pairwise human preferences in LLM responses.\\

\textbf{Direct Preference Optimisation.} Although RLHF has proven effective in homogeneous and heterogeneous preference settings (Park. et. al. \cite{hetero_rlhf}), it is complex and faces instability issues (Casper et. al. \cite{comp_rlhf}) during training. To overcome this, Rafailov et. al. \cite{dpo} proposed Direct Preference Optimisation that performed preference optimisation a single step instead. DPO provided several computational and stability advantages over RLHF (Liu et. al. \cite{dpo_adv}) that ensured its popularity in the community. As a result, multiple DPO variants have emerged (Zeng et. al. \cite{tdpo}, Lanchantin et. al. \cite{divpo}, Shao et. al. \cite{grpo}, Zhou et. al. \cite{modpo}) that improve upon certain aspects of the original DPO algorithm.

\textbf{2-Dimensional DPO.} Prefernce optimisation algorithms as a whole operate by assigning a single score to the entire LLM response that shall distinguish the preferred and rejected responses. This assumption inhibits the algorithms to capture true ``granularity'' of human preferences. Human preference inherently have multiple \textbf{aspects} (Stokes et. al. \cite{div_pref}) that are not captured by regular preference optimisation models. Li et. al. \cite{2ddpo} proposed the 2-dimension DPO (2D-DPO) which would score each segment of the response across 5 segments - \textit{Completeness, Clarity, Correctness, Safety} \& \textit{Helpfulness}. A combination of these scores then create a more holistic view of human preferences and enable more accuracy preference optimisation in LLMs. \\

\textbf{Robust DPO.} Preference optimisation algorithms also operate under another major assumption which is that human annotations are the ground truth. Practically, this assumption doesn't always hold true as there can be noise in human labels and scores (Wei et. al. \cite{noisy_labels}). Specifically, noisy (incorrect and ambiguous) preference pairs in the dataset might restrict the language models from capturing human intent accurately. While practitioners have recently proposed heuristics to mitigate the effect of noisy preferences, a complete theoretical understanding of their workings remained elusive till Chowdhury et. al. \cite{robdpo} proposed Robust DPO - a general framework for policy optimisation in the presence of random preference flips. This was achieved by using a novel loss function, which de-biases the effect of noise on average, making a policy trained by minimizing a loss robust to the noise. Although this was performed for Vanilla DPO, there is a need to design a similar framework for other DPO techniques. Our work aims to add a sense of robustness similar to the work by Chowdhury et. al. \cite{robdpo} in a 2D-DPO setting.

\section{Problem Setup and Background}
Preference optimisation algorithms take in preference dataset $\mathcal{D} \hspace{0.1cm} = \hspace{0.1cm} {\left\{ {s_i}, {a_{w}^{i}}, {a_{l}^{i}} \right\}}_{i = 1}^{n}$ where ${s_i} \sim \rho$ is a prompt sampled from the space of possible prompts and ${a_{w}^{i}}, {a_{l}^{i}}$ denote the preferred and rejected responses respectively. Assuming a BTL model (\cite{btl}) and a latent reward model ${r ^ \ast} \hspace{0.1cm} \colon \hspace{0.1cm} \left( s, a \right) \rightarrow \mathbb{R}$, RLHF (\cite{rlhf}) aims to perform the following 2 steps - 
\begin{enumerate}
    \item It learns an optimal reward model $r ^ \ast$ such that the margin between the scores for $a_w$ and $a_l$ is maximized across prompts and responses.
    \item Then, it uses $r ^ \ast$ to find an optimal policy $\pi ^ \ast$ that maximizes the probability of getting $a_{w}^{i}$ for prompt $s^i$ across all $i$
\end{enumerate}

This formulation for RLHF results in the Eq. \ref{eq:3.1} which gives the optimisation problem for getting the optimal policy - 
\begin{equation}\label{eq:3.1}
    {\pi ^ \ast} \hspace{0.1cm} = \hspace{0.1cm} \arg \max_\pi \hspace{0.1cm} J(\pi) \hspace{0.1cm} = \hspace{0.1cm} \arg \max_\pi \hspace{0.1cm} \mathbb{E}_{s \sim \rho, \hspace{0.05cm} a \sim \pi(. \hspace{0.05cm} \vert \hspace{0.05cm} s)} \hspace{0.1cm} \left[ {r ^ \ast}(s, a) \hspace{0.1cm} - \hspace{0.1cm} \beta \log \left( \frac{\pi(a | s)}{{\pi ^ \text{SFT}} (a | s)} \right)\right]
\end{equation}

This follows from the usage of the BTL model (\cite{btl}) where the probability of choosing the preferred response over the rejected response for a prompt $s$ is given in Eq. \ref{eq:3.2} 
\begin{equation}\label{eq:3.2}
    \mathbb{P} \left( {a_w} \succ {a_l} \hspace{0.1cm} \vert \hspace{0.1cm} s \right) \hspace{0.1cm} = \hspace{0.1cm} \sigma \left[ {r ^ \ast}(s, {a_w}) \hspace{0.1cm} - \hspace{0.1cm} {r ^ \ast}(s, {a_l}) \right]
\end{equation}

Rafailov et. al. \cite{dpo} proposed \textbf{Direct Preference Optimisation} that would write the reward model in terms of the policy and directly optimize over the policy in a single step. This formulation was more stable and resulted in the optimal policy as mentioned in Eq. \ref{eq:3.3}.

\begin{equation}\label{eq:3.3}
    \begin{split}
        {\pi ^ \ast} & \hspace{0.1cm} = \hspace{0.1cm} \arg \max_\theta \hspace{0.1cm} \mathcal{L} \left( \theta \hspace{0.05cm} ; \hspace{0.05cm} s, {a_w}, {a_l} \right) \hspace{0.1cm} \\
        & = \hspace{0.1cm} - \log \sigma \left[ \beta \log \left( \frac{{\pi_\theta}({a_w} | s)}{{\pi ^ \text{SFT}} ({a_w} | s)} \right) \hspace{0.1cm} - \hspace{0.1cm} \beta \log \left( \frac{{\pi_\theta}({a_l} | s)}{{\pi ^ \text{SFT}} ({a_l} | s)} \right) \right] \\
        & = \hspace{0.1cm} - \log \sigma \left( \beta {h_\theta} \left( s, {a_w}, {a_l} \right) \right)
    \end{split}
\end{equation}

Where ${h_\theta} \left( s, {a_w}, {a_l} \right) \hspace{0.1cm} = \hspace{0.1cm} {r_\theta} (s, {a_w}) - {r_\theta}(s, {a_l})$. The mathematical proof for the above expression is provided in Appendix \ref{app:a}. \\

Eq. \ref{eq:3.3} is the vanilla DPO loss formulation. However, this formulation is not robust to label noise in the dataset. Chowdhury et. al. \cite{robdpo} proposed a more robust formulation of DPO in which noise was induced in the preferences via the standard random noise model, where the revealed preferences are true preferences flipped with a small probability $\epsilon \in \left(0, \frac{1}{2} \right)$ i.e. 
\begin{equation}\label{eq:3.4}
    {\mathbb{P}_\epsilon} \left[ \left( \Tilde{a_w}, \Tilde{a_l} \right) \hspace{0.1cm} = \hspace{0.1cm} \left( {a_l}, {a_w} \right) \hspace{0.1cm} \vert \hspace{0.1cm} s \right] \hspace{0.1cm} = \hspace{0.1cm} \epsilon
\end{equation}

Similar to Vanilla DPO, the loss function for the noisy dataset can be written as seen in Eq. \ref{eq:3.5}.
\begin{equation}\label{eq:3.5}
    \mathcal{L_\epsilon} \left( \theta \hspace{0.05cm} ; \hspace{0.05cm} s, \Tilde{a_w}, \Tilde{a_l} \right) \hspace{0.1cm} = \hspace{0.1cm} - \log {\mathbb{P}_{\theta, \epsilon}} \left[ \Tilde{a_w} \succ \Tilde{a_l} \hspace{0.1cm} \vert \hspace{0.1cm} s \right]
\end{equation}

However, this formulation has a major drawback as it introduces a \textbf{bias} in the DPO loss. This is because - 
\begin{equation}\label{eq:3.6}
    \mathrm{logit} \left( {\mathbb{P}_\theta} \left[ {a_w} \succ {a_l} \hspace{0.1cm} \vert \hspace{0.1cm} s \right] \right) \hspace{0.1cm} \ne \hspace{0.1cm} \mathrm{logit} \left( {\mathbb{P}_{\theta, \epsilon}} \left[ {a_w} \succ {a_l} \hspace{0.1cm} \vert \hspace{0.1cm} s \right] \right)
\end{equation}
Thus, there is a need to modify the formulation to take care of the bias. An unbiased loss formulation was presented by Chowdhury et. al. \cite{robdpo} as shown in \ref{eq:3.7}.
\begin{equation}\label{eq:3.7}
    {\hat{\mathcal{L}}_{\epsilon}} \left( \theta \hspace{0.1cm} ; \hspace{0.1cm} s, {\Tilde{a}_w}, {\Tilde{a}_l} \right) \hspace{0.1cm} = \hspace{0.1cm} \frac{\left( 1 - \epsilon \right) \mathcal{L}\left( \theta \hspace{0.1cm} ; \hspace{0.1cm} s, {\Tilde{a}_w}, {\Tilde{a}_l} \right) \hspace{0.1cm} - \hspace{0.1cm} \epsilon \mathcal{L} \left( \theta \hspace{0.1cm} ; \hspace{0.1cm} s, {\Tilde{a}_l}, {\Tilde{a}_2} \right)}{1 - 2 \epsilon}
\end{equation}

A detailed derivation and mathematical proof as to why this loss is unbiased is presented in \ref{app:b}

\section{Methodology}
In the previous section, we explored a robust model to take care of noisy labels in Vanilla DPO (when $\epsilon$ is small) as proposed by Chowdhury et. al. \cite{robdpo}. Although robust, this method still doesn't enjoy the granularity that 2D-DPO offers and hence, a clear need to extend robustness to the 2D-DPO paradigm has been raised. The algorithm proposed by us in this work aims to induce robustness in the 2D-DPO algorithm (as proposed by Li et. al. \cite{2ddpo}) at a \textbf{segment level}.

\subsection{Understanding 2D-DPO}

In order to induce robustness in 2D-DPO, we first need to explore the Vanilla 2D-DPO algorithm as proposed by Li et. al. \cite{2ddpo}. 2D-DPO takes a prompt as an input and each prompt has a preferred response $\left( {\tau_w} \right)$ and a rejected response $\left( {\tau_l} \right)$. These responses are further divided into segments by being split over grammatical sentence separators (like periods, commas, etc.). Consider (without a loss of generality) that the winner (preferred) response $\tau_w$ has $N_w$ number of segments after splitting and similarly, the loser (rejected) response $\tau_l$ has $N_l$ segments. As $N_w$ and $N_l$ need not be equal, we take inspiration from \cite{2ddpo} and define $$ N = \min \left( {N_w}, {N_l} \right)$$ Then, we proceed to choose the top-$N$ segments of $\tau_w$ while choosing the bottom-$N$ segments of $\tau_l$ \footnote{Choosing the top-$N$ segments from the winner response and bottom-$N$ segments from the loser response is done to maximize the margin between the two.}\\

The segments of the responses are indexed by $k \in \{0, 1, \hdots , N-1\}$ and for the $k^{\mathrm{th}}$ segment, token-level DPO (as proposed by Zeng et. al. \cite{tdpo}) is performed to calculate the probability of getting token $a_t$ given token $s_t$ has arrived. Easch of the segments is further scored based on \textbf{5 aspects} - ``Completeness'', ``Clarity'', ``Correctness'', ``Safety'' and ``Helpfulness''. These scores are assigned to each segment by external annotators \footnote{In the work proposed by \cite{2ddpo}, GPT-4 Turbo is used to simulate external annotators and provide the aspect scores. Check Appendix F of \cite{2ddpo} for more details}. Each segment is given an integer score between 0 and 4 (both inclusive) across an aspect where score 4 reflects high desirability and score 0 reflects undesirability.\\

These aspect scores are then combined in a weighted combination to get the score of the segment as shown in Eq. \ref{eq:4.1}

\begin{equation}\label{eq:4.1}
    \begin{split}
        {r_{w,k}} \hspace{0.1cm} = \hspace{0.1cm} {w^T}{r_k} \hspace{0.1cm} & = \hspace{0.1cm} \left( {w_{\mathrm{completeness}}} \hspace{0.05cm} \ast \hspace{0.05cm} {r_{\mathrm{completeness}}} \right) \hspace{0.1cm} + \hspace{0.1cm} \left( {w_{\mathrm{clarity}}} \hspace{0.05cm} \ast \hspace{0.05cm} {r_{\mathrm{clarity}}} \right) \hspace{0.1cm} + \hspace{0.1cm} \left( {w_{\mathrm{correctness}}} \hspace{0.05cm} \ast \hspace{0.05cm} {r_{\mathrm{correctness}}} \right) \\
        & + \hspace{0.1cm} \left( {w_{\mathrm{safety}}} \hspace{0.05cm} \ast \hspace{0.05cm} {r_{\mathrm{safety}}} \right) \hspace{0.1cm} + \hspace{0.1cm} \left( {w_{\mathrm{helpfulness}}} \hspace{0.05cm} \ast \hspace{0.05cm} {r_{\mathrm{helpfulness}}} \right)
    \end{split}
\end{equation}

In Eq. \ref{eq:4.1}, ${r_{w,k}}$ represents the score for the $k^\mathrm{th}$ segment of the winner response $\tau_w$ and this is a weighted combination of the scores this segment has gotten across the aspects. One thing to note here is that $$ {\sum_{i = 1}^5} \hspace{0.1cm} {w_i} \hspace{0.1cm} = \hspace{0.1cm} 1 \hspace{0.5cm} ; \hspace{0.5cm} w_i \geq 0 \hspace{0.2cm} \forall \hspace{0.1cm} i $$

Therefore, the segment score is a convex combination of the aspect scores for that segment and thus ${r_{w,k}} \in [0,4]$ i.e. the segment scores are real numbers between 0 and 4 (both inclusive). This definition of the segment and aspect scores can now be used to formulate the loss function for the 2D-DPO paradigm as illustrated by \cite{2ddpo}.

\begin{equation}\label{eq:4.2}
    \begin{split}
        {\mathcal{L}_\mathrm{group}} \left( {\pi_\theta} ; \mathcal{D} \right) \hspace{0.1cm} & = \hspace{0.1cm} - {\mathbb{E}_{\left( {\tau_w}, {\tau_l} \right) \sim \mathcal{D}}} [ {\sum_{k = 0}^{N - 1}} \hspace{0.05cm} \log \sigma \{ \beta \hspace{0.05cm} {r_{w,k}}{\sum_{t = {n_k}}^{{n_k} + {l_k}}} \hspace{0.05cm}  \log \frac{{\pi_\theta} \left( {a_{w}^{t}} \hspace{0.05cm} \vert \hspace{0.05cm} {s_{w}^{t}} \right)}{{\pi_\mathrm{ref}} \left( {a_{w}^{t}} \hspace{0.05cm} \vert \hspace{0.05cm} {s_{w}^{t}} \right)} \\
            & - \hspace{0.1cm} \beta \hspace{0.05cm} {{r}_{l,k}} {\sum_{t = {n_k}}^{{n_k} + {l_k}}} \hspace{0.05cm} \log \frac{{\pi_\theta} \left( {a_{l}^{t}} \hspace{0.05cm} \vert \hspace{0.05cm} {s_{l}^{t}} \right)}{{\pi_\mathrm{ref}} \left( {a_{l}^{t}} \hspace{0.05cm} \vert \hspace{0.05cm} {s_{l}^{t}} \right)} \} ]
    \end{split}
\end{equation}

Comparing Eq. \ref{eq:4.2} with \ref{eq:3.3}, we can see that there is a similarity with the original DPO paradigm. The intuition to maximize the difference between the token log-probalities of the winner and loser responses still holds in Eq. \ref{eq:4.2}. However, now each of the log-probability values is being calculated for a segment and then weighted with the segment scores. Then, these differences are fed to the log-sigmoid function to get the preference log-probabilities as per the \textbf{BTL - model} (\cite{btl}). This would give us the log-probabilities of the $k^\mathrm{th}$ segments of the preferred and rejected responses. The differences are then summed over all segments and the expectation over the responses for this quantity completes the loss function.

\subsection{High-Level Noise-Robust 2D-DPO Formulation}

Despite the loss formulation in Eq. \ref{eq:4.2} being more fine-grained when compared to the Vanilla DPO loss function, it is not robust to noise in the scores. Since the scores for each segment across aspects are given by external annotators, these scores may have some level of label noise, and a robust formulation is needed to handle this noise. As per the formulation in Eq. \ref{eq:4.2}, there can be multiple places where noise could be induced in the model. \\

The most high-level noise could take the form of \textbf{``preference flips''} i.e. the annotator labels the preferred response as the rejected response and vice-versa with some probability $\gamma$. Mathematically, this would mean 
\[
    {r_{w,k}}{\sum_{t = {n_k}}^{{n_k} + {l_k}}} \hspace{0.05cm}  \log \frac{{\pi_\theta} \left( {a_{w}^{t}} \hspace{0.05cm} \vert \hspace{0.05cm} {s_{w}^{t}} \right)}{{\pi_\mathrm{ref}} \left( {a_{w}^{t}} \hspace{0.05cm} \vert \hspace{0.05cm} {s_{w}^{t}} \right)} \hspace{0.3cm} \Longleftrightarrow \hspace{0.3cm} {{r}_{l,k}} {\sum_{t = {n_k}}^{{n_k} + {l_k}}} \hspace{0.05cm} \log \frac{{\pi_\theta} \left( {a_{l}^{t}} \hspace{0.05cm} \vert \hspace{0.05cm} {s_{l}^{t}} \right)}{{\pi_\mathrm{ref}} \left( {a_{l}^{t}} \hspace{0.05cm} \vert \hspace{0.05cm} {s_{l}^{t}} \right)} \hspace{0.5cm} \text{with probability $\gamma$}
\]

In simpler terms, the sign of the quantity inside the sigmoid in Eq. \ref{eq:4.2} will be flipped. If the sum of the token-level log-probabilities weighted by the segment rewards evaluate to \textbf{0}, then Vanilla 2D-DPO shall remain robust to this noise model. This follows from Lemma \ref{lem:4.1}

\begin{lemma}
    \label{lem:4.1}
    For any $x \in \mathbb{R}$ and the $\sigma(x)$ begin defined as the standard Sigmoid function, we have - 
    \begin{equation}
        \label{eq:4.16}
        \log \left[ \sigma (x) \right] =   \log \left[ \sigma (-x) \right] \hspace{0.3cm} \Longleftrightarrow \hspace{0.3cm} x = 0
    \end{equation}
\end{lemma}

Proof of Lemma \ref{lem:4.1} has been deferred to Appendix \ref{app:c}. \\

If one notices carefully, then this noise model is almost the same one that Chowdhury et. al. \cite{robdpo} consider in their work and thus, the unbiased loss estimator in Eq. \ref{eq:3.7} works for our use case as well. However, the loss $\mathcal{L}\left( \theta \hspace{0.1cm} ; \hspace{0.1cm} s, {\Tilde{a}_w}, {\Tilde{a}_l} \right)$ will now be taken in from the 2D-DPO loss formulation in Eq. \ref{eq:4.2}. The noisy-robust unbiased loss estimator in this case can be given as shown in Eq. \ref{eq:4.17}

\begin{equation}
    \label{eq:4.17}
    {\hat{\mathcal{L}}_{\gamma}} \left( \pi_\theta \hspace{0.1cm} ; \hspace{0.1cm} \mathcal{D} \right) \hspace{0.1cm} = \hspace{0.1cm} \frac{\left( 1 - \gamma \right) {\mathcal{L}_\mathrm{group}} \left( {\pi_\theta} ; \mathcal{D} \right) \hspace{0.1cm} - \hspace{0.1cm} \gamma {\mathcal{L}_\mathrm{group}} \left( {\pi_\theta} ; \mathcal{D} \right)}{1 - 2 \gamma}
\end{equation}

\subsection{Segment-Level Noise-Robust 2D-DPO Formulation}
The loss mentioned in the previous section is a high-level noise where entire response preferences are flipped with some probability. However, this flipping is not practically observed, unlike \textbf{segment-level} noise. In this case, there is a small perturbation $\delta$ added to the segment score $r_{w,k}$ or $r_{l,k}$. This perturbation is uniformly sampled from $[0,1]$ since having a perturbation beyond 1 can induce a large amount of error that might not be manageable. Let us consider the loss function of 2D-DPO as written in Eq. \ref{eq:4.2} - 
\begin{align*}
    {\mathcal{L}_\mathrm{group}} \left( {\pi_\theta} ; \mathcal{D} \right) \hspace{0.1cm} & = \hspace{0.1cm} - {\mathbb{E}_{\left( {\tau_w}, {\tau_l} \right) \sim \mathcal{D}}} [ {\sum_{k = 0}^{N - 1}} \hspace{0.05cm} \log \sigma \{ \beta {\sum_{t = {n_k}}^{{n_k} + {l_k}}} \hspace{0.05cm} {r_{w,k}} \log \frac{{\pi_\theta} \left( {a_{w}^{t}} \hspace{0.05cm} \vert \hspace{0.05cm} {s_{w}^{t}} \right)}{{\pi_\mathrm{ref}} \left( {a_{w}^{t}} \hspace{0.05cm} \vert \hspace{0.05cm} {s_{w}^{t}} \right)} \\
        & - \hspace{0.1cm} \beta {\sum_{t = {n_k}}^{{n_k} + {l_k}}} \hspace{0.05cm} {{r}_{l,k}} \log \frac{{\pi_\theta} \left( {a_{l}^{t}} \hspace{0.05cm} \vert \hspace{0.05cm} {s_{l}^{t}} \right)}{{\pi_\mathrm{ref}} \left( {a_{l}^{t}} \hspace{0.05cm} \vert \hspace{0.05cm} {s_{l}^{t}} \right)} \} ]
\end{align*}

To simplify the notation, let us define the following terms.
\begin{equation}
    \label{eq:4.3}
    {l_{(w,k)}} \hspace{0.1cm} = \hspace{0.1cm} \beta {\sum_{t = {n_k}}^{{n_k} + {l_k}}} \hspace{0.05cm} \log \frac{{\pi_\theta} \left( {a_{w}^{t}} \hspace{0.05cm} \vert \hspace{0.05cm} {s_{w}^{t}} \right)}{{\pi_\mathrm{ref}} \left( {a_{w}^{t}} \hspace{0.05cm} \vert \hspace{0.05cm} {s_{w}^{t}} \right)} \hspace{0.5cm} ; \hspace{0.5cm} {l_{(l,k)}} \hspace{0.1cm} = \hspace{0.1cm} \beta {\sum_{t = {n_k}}^{{n_k} + {l_k}}} \hspace{0.05cm} \log \frac{{\pi_\theta} \left( {a_{l}^{t}} \hspace{0.05cm} \vert \hspace{0.05cm} {s_{l}^{t}} \right)}{{\pi_\mathrm{ref}} \left( {a_{l}^{t}} \hspace{0.05cm} \vert \hspace{0.05cm} {s_{l}^{t}} \right)}
\end{equation}

Substituting from Eq. \ref{eq:4.3} back into Eq. \ref{eq:4.2}, we get - 
\begin{equation}
    \label{eq:4.4}
    {\mathcal{L}_\mathrm{group}} \left( {\pi_\theta} ; \mathcal{D} \right) \hspace{0.1cm} = \hspace{0.1cm} - {\mathbb{E}_{\left( {\tau_w}, {\tau_l} \right) \sim \mathcal{D}}} \hspace{0.1cm} \left[ {\sum_{k = 0}^{N - 1}} \hspace{0.05cm} \log \sigma \left\{ {r_{(w,k)}} {l_{(w,k)}} \hspace{0.1cm} - \hspace{0.1cm} {r_{(l,k)}} {l_{(l,k)}} \right\} \right]
\end{equation}

Here, ${r_{(w,k)}}$ and ${r_{(l,k)}}$ represent the segment scores for a segment $k$ of the preferred and rejected responses respectively. To these scores, let us now add a small perturbation $\delta$ to get the noisy scores as seen in Eq. \ref{eq:4.5}

\begin{equation}
    \label{eq:4.5}
    {\hat{r}_{(w,k)}} \hspace{0.1cm} = \hspace{0.1cm} {r_{(w,k)}} - \delta \hspace{0.5cm} ; \hspace{0.5cm} {\hat{r}_{(l,k)}} \hspace{0.1cm} = \hspace{0.1cm} {r_{(l,k)}} + \delta
\end{equation}

The perturbation shall lower the preferred response's segment scores while it shall increase the loser response's segment scores. This noise model is aimed to reduce the margin between the 2 scores and hence induce a level of confusion in the model. Substituting the noisy rewards from Eq. \ref{eq:4.5} back into the loss function as defined in Eq. \ref{eq:4.4}, we get the noisy loss function - 
\begin{equation}
    \label{eq:4.6}
    \begin{split}
        \hat{L}({\pi_\theta}, \mathcal{D}, \delta) \hspace{0.1cm} & = \hspace{0.1cm} - {\mathbb{E}_{\delta \sim U(0,1)}}{\mathbb{E}_{\left( {\tau_w}, {\tau_l} \right) \sim \mathcal{D}}} \hspace{0.1cm} \left[ {\sum_{k = 0}^{N - 1}} \hspace{0.05cm} \log \sigma \left\{ {\hat{r}_{(w,k)}} {l_{(w,k)}} \hspace{0.1cm} - \hspace{0.1cm} {\hat{r}_{(l,k)}} {l_{(l,k)}} \right\} \right] \\
        \hspace{0.1cm} & = \hspace{0.1cm} - {\mathbb{E}_{\delta \sim U(0,1)}}{\mathbb{E}_{\left( {\tau_w}, {\tau_l} \right) \sim \mathcal{D}}} \hspace{0.1cm} \left[ {\sum_{k = 0}^{N - 1}} \hspace{0.05cm} \log \sigma \left\{ \left( {r_{(w,k)}} - \delta \right) {l_{(w,k)}} \hspace{0.1cm} - \hspace{0.1cm} \left( {r_{(l,k)}} + \delta \right) {l_{(l,k)}} \right\} \right] \\
        \hspace{0.1cm} & = \hspace{0.1cm} - {\mathbb{E}_{\delta \sim U(0,1)}}{\mathbb{E}_{\left( {\tau_w}, {\tau_l} \right) \sim \mathcal{D}}} \hspace{0.1cm} \left[ {\sum_{k = 0}^{N - 1}} \hspace{0.05cm} \log \sigma \left\{{r_{(w,k)}} {l_{(w,k)}} \hspace{0.1cm} - \hspace{0.1cm} {r_{(l,k)}} {l_{(l,k)}} - \delta \left( {l_{(w,k)}} + {l_{(l,k)}} \right) \right\} \right] \\
        \hspace{0.1cm} & = \hspace{0.1cm} - {\mathbb{E}_{\delta \sim U(0,1)}}{\mathbb{E}_{\left( {\tau_w}, {\tau_l} \right) \sim \mathcal{D}}} \hspace{0.1cm} \left[ {\sum_{k = 0}^{N - 1}} \hspace{0.05cm} \log \sigma \left\{{X_k} - \delta {Y_k} \right\} \right]
    \end{split}
\end{equation}

Where
\begin{equation}
    \label{eq:4.7}
    {X_k} \hspace{0.1cm} = \hspace{0.1cm} {r_{(w,k)}} {l_{(w,k)}} \hspace{0.1cm} - \hspace{0.1cm} {r_{(l,k)}} {l_{(l,k)}} \hspace{0.5cm} ; \hspace{0.5cm} {Y_k} \hspace{0.1cm} = \hspace{0.1cm} {l_{(w,k)}} + {l_{(l,k)}}
\end{equation}

Thus, our optimisation problem has now become - 
\begin{equation}
    \label{eq:4.8}
    \begin{split}
        \text{minimize} \hspace{0.5cm} &  - {\mathbb{E}_{\delta \sim U(0,1)}} {\mathbb{E}_{\left( {\tau_w}, {\tau_l} \right) \sim \mathcal{D}}} \hspace{0.1cm} \left[ {\sum_{k = 0}^{N - 1}} \hspace{0.05cm} \log \sigma \left\{{X_k} - \delta {Y_k} \right\} \right]
    \end{split}
\end{equation}

The optimisation problem in Eq. \ref{eq:4.8} is unconstrained in $\theta$ while being constrained between $[0,1]$ in $\delta$ and doesn't have a closed form solution. Such loss functions are solved using \textbf{Gradient-based} methods. We will be using Stochastic Gradient Descent in this work, as shown in the algorithm in Appendix \ref{app:d}.

\section{Experiments}\footnote{Code and Experiment details (to replicate) can be found \href{https://github.com/nachiketapatil/Robust-2D-DPO}{here}}
We performed experiments to first create a base line and then compare our algorithm against said baseline. Tab \ref{tab:exps} contains the list of experiments performed in this work. The model used in all the experiments is \textbf{Pythia 6.9B}.

\begin{table}[h]
    \centering
    \begin{tabular}{|p{2cm}|p{2cm}|p{3cm}|p{3cm}|p{3cm}|}
        \hline
        \textbf{Experiment Number} & \textbf{Algorithm} & \textbf{Dataset (train)} & \textbf{Dataset (Eval)} & \textbf{Purpose} \\
        \hline
        1 & DPO & HelpSteer-2D (Original) & HelpSteer-2D (Original) & To set a baseline. \\
        \hline
        2 & 2D-DPO & HelpSteer-2D (Original) & HelpSteer-2D (Original) & To compare against DPO. \\
        \hline
        3 & 2D-DPO & HelpSteer-2D (Original) & HelpSteer-2D (Noisy) & To notice effect of noise on win-rates. \\
        \hline
        4 & 2D-DPO & HelpSteer-2D (Noisy) & HelpSteer-2D (Noisy) & To check the functioning of our formulation. \\
        \hline
        5 & DPO & Anthropic-HH & Anthropic-HH & For a sanity check on DPO code \\
        \hline
    \end{tabular}
    \caption{Experiments Performed}
    \label{tab:exps}
\end{table}

Since Experiment 5 is performed for a sanity check, the details have been pushed to Appendix \ref{app:e}.

\subsection{Experiment 1}
\begin{figure}[h!]
    \centering
    \begin{subfigure}[t]{0.52\textwidth}
        \centering
        \includegraphics[width =\linewidth]{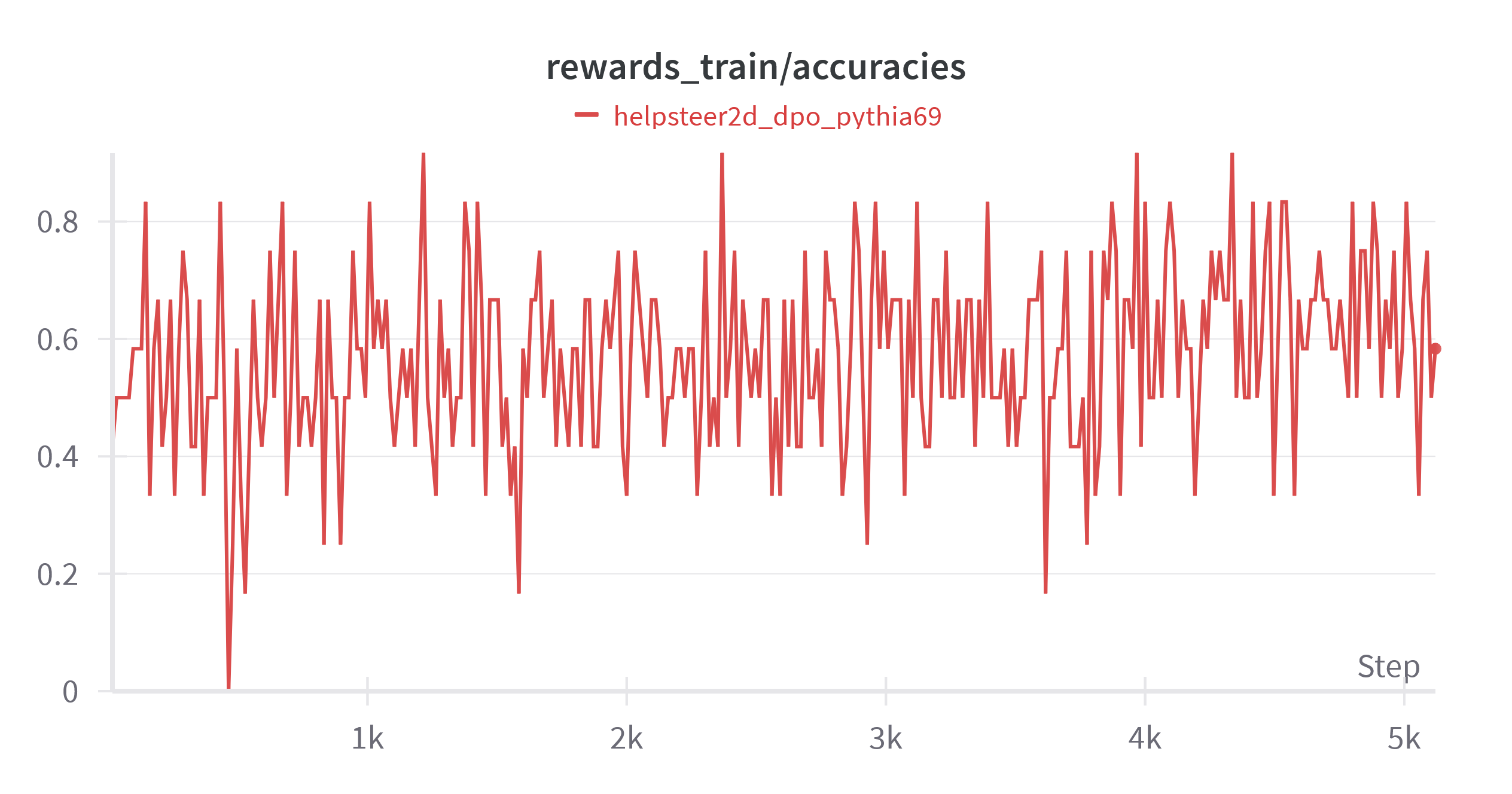}
        \caption{During Training}
    \end{subfigure}%
    ~ 
    \begin{subfigure}[t]{0.52\textwidth}
        \centering
        \includegraphics[width = \linewidth]{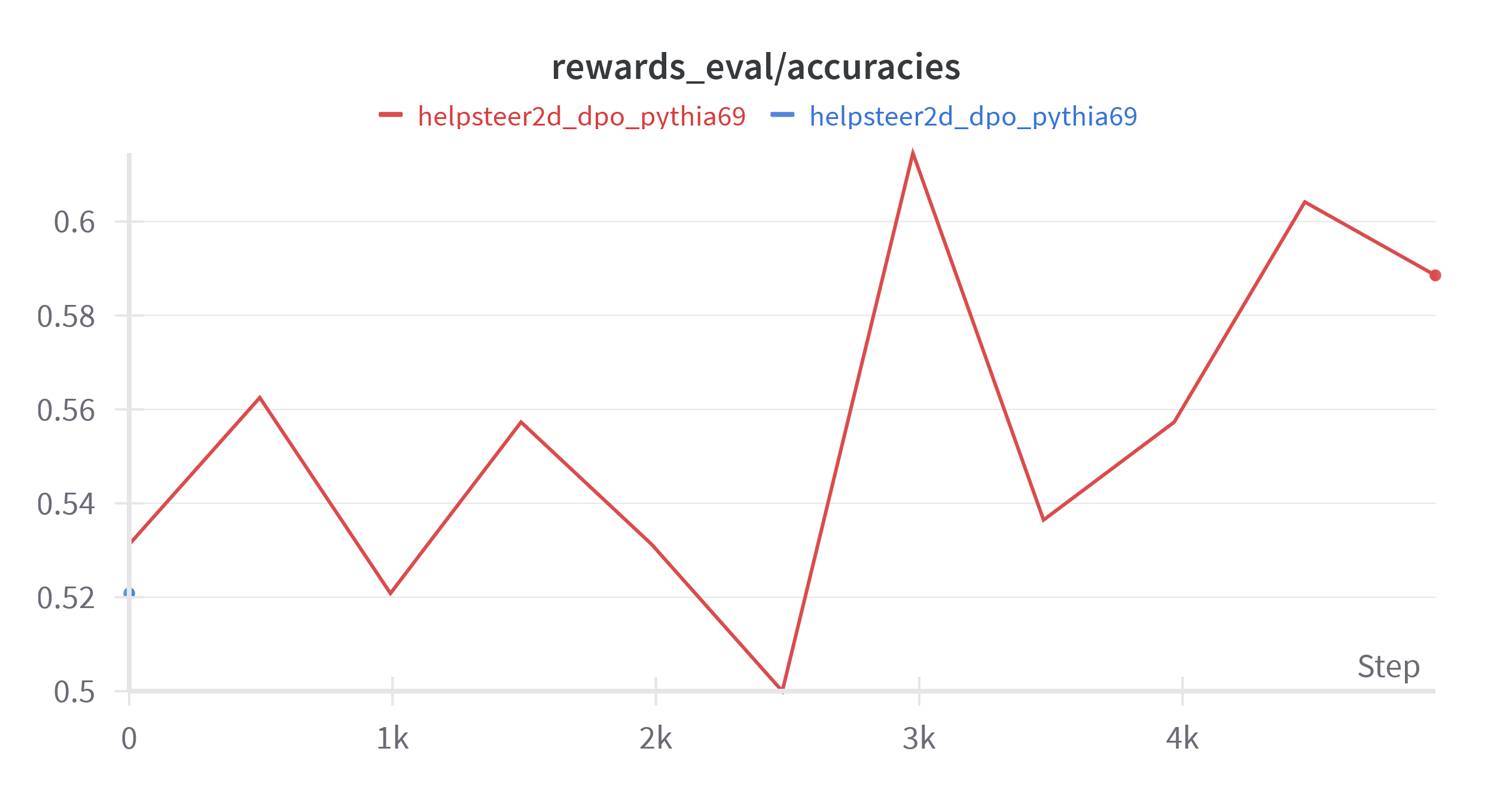}
        \caption{During Evaluation}
    \end{subfigure}%
    \caption{Win rate progression for Vanilla DPO}
    \label{fig:win_rate_1d}
\end{figure}

In this experiment, the Vanilla DPO algorithm (\cite{dpo}) was executed on the \verb|HelpSteer-2D| dataset. The dataset contains \textbf{6400} prompts and each prompt has a chosen and rejected response pair. Each of the responses also has an associated array of scores between $0$ and $4$ inclusive. These are the scores for each of the segments of the response. Fig. \ref{fig:win_rate_1d} shows the progression of win-rate during training and evaluation.

The final win-rate achieved during training was \textbf{58.333\%} and during evaluation was \textbf{58.854\%}

\subsection{Experiment 2}
\begin{figure}[h!]
    \centering
    \begin{subfigure}[t]{0.52\textwidth}
        \centering
        \includegraphics[width = \linewidth]{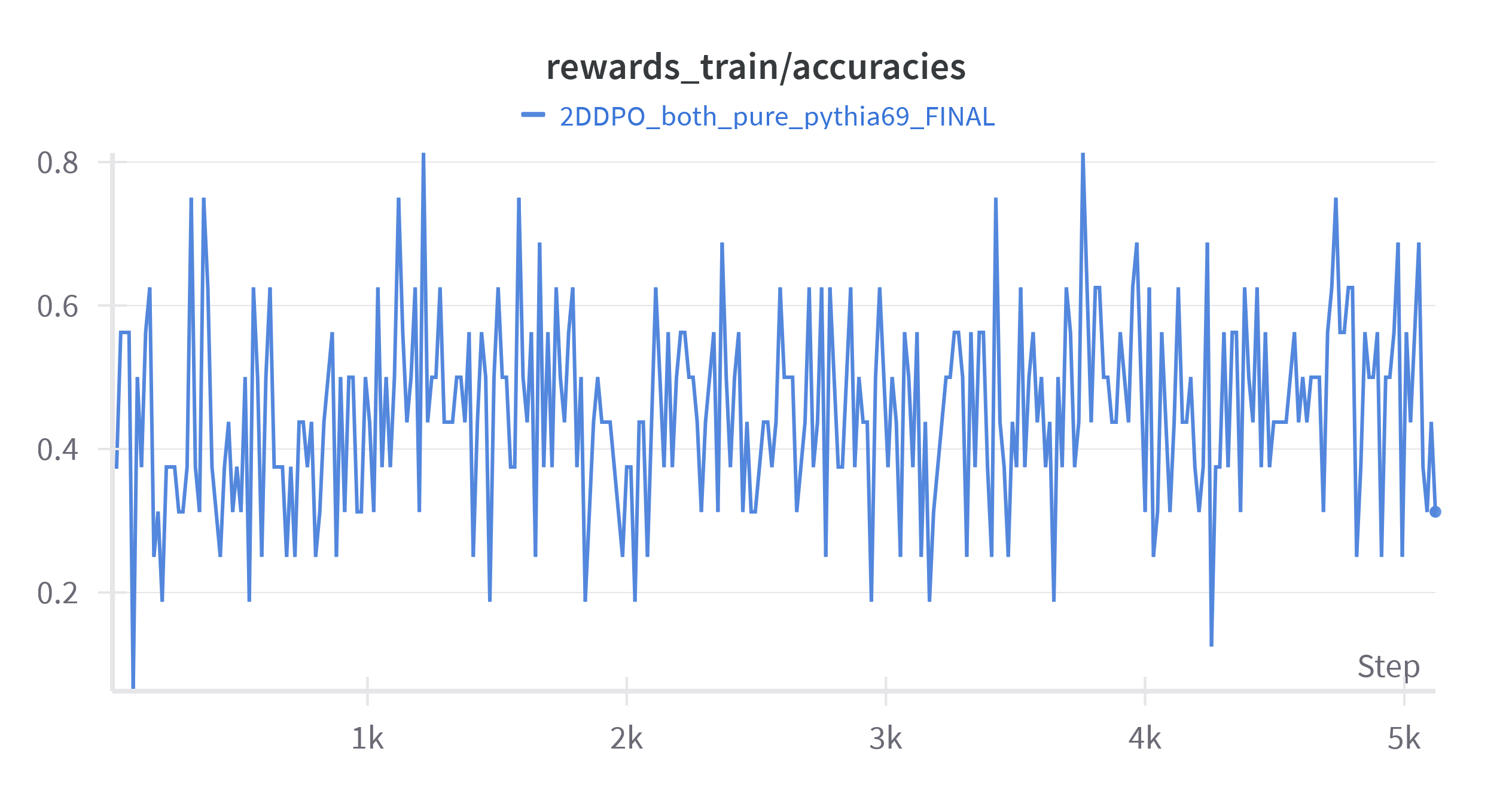}
        \caption{During Training}
    \end{subfigure}%
    ~ 
    \begin{subfigure}[t]{0.52\textwidth}
        \centering
        \includegraphics[width = \linewidth]{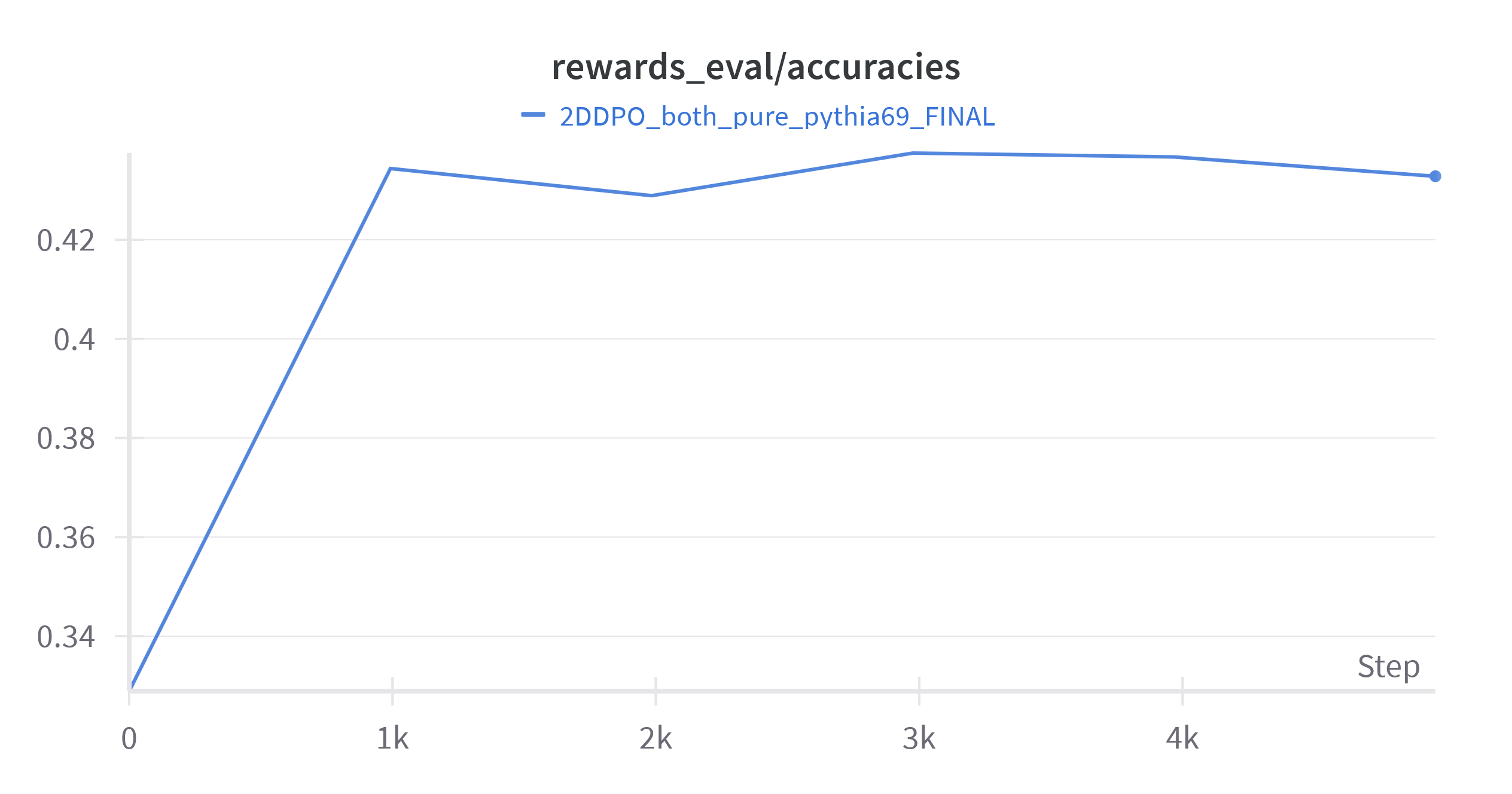}
        \caption{During Evaluation}
    \end{subfigure}%
    \caption{Win rate progression for Vanilla 2D-DPO (Noiseless)}
    \label{fig:win_rate_2d}
\end{figure}

In this experiment, the Vanilla 2D-DPO algorithm (\cite{2ddpo}) was executed on the \verb|HelpSteer-2D| dataset. No segment level noise had been introduced yet. Fig. \ref{fig:win_rate_2d} shows the progression of win-rate during training and evaluation.

The final win-rate achieved during training was \textbf{31.25\%} and during evaluation was \textbf{43.281\%}

\subsection{Experiment 3}
\begin{figure}[h!]
    \centering
    \begin{subfigure}[t]{0.52\textwidth}
        \centering
        \includegraphics[width = \linewidth]{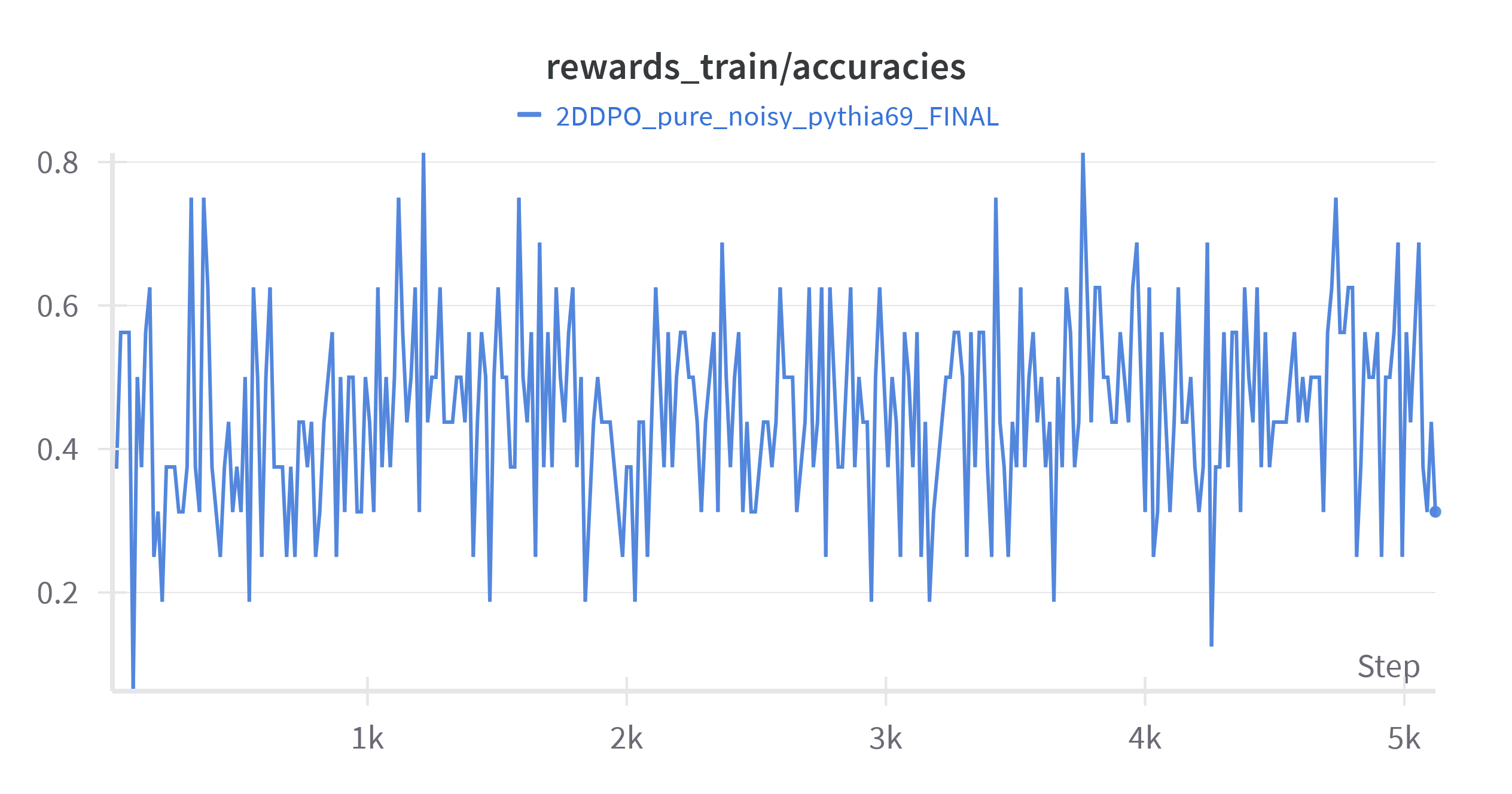}
        \caption{During Training}
    \end{subfigure}%
    ~ 
    \begin{subfigure}[t]{0.52\textwidth}
        \centering
        \includegraphics[width = \linewidth]{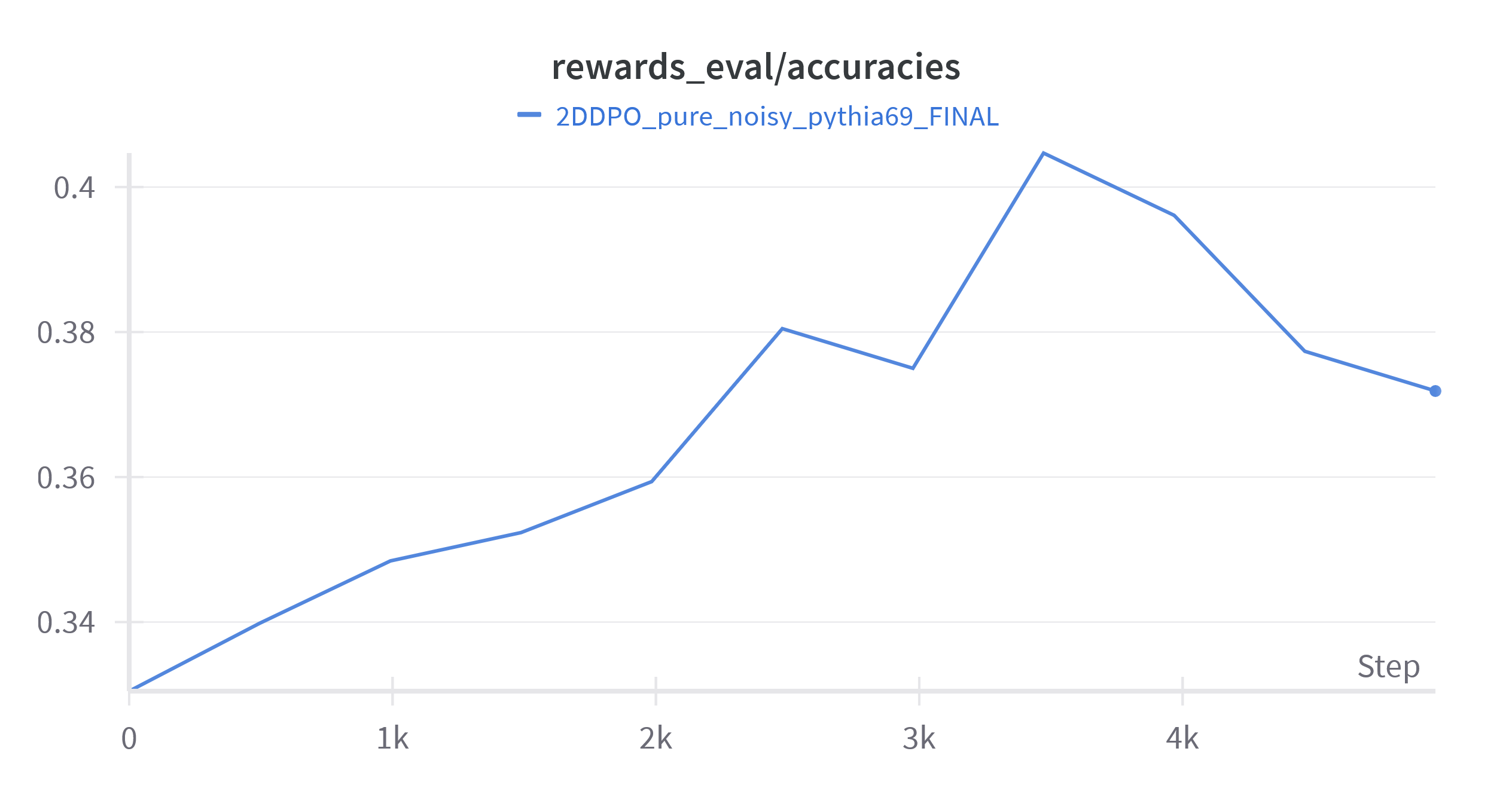}
        \caption{During Evaluation (Noisy)}
    \end{subfigure}%
    \caption{Win rate progression for Vanilla 2D-DPO}
    \label{fig:win_rate_2d_cn}
\end{figure}

In this experiment, the Vanilla 2D-DPO algorithm (\cite{2ddpo}) was executed on the \verb|HelpSteer-2D| dataset. No segment level noise has been introduced during training but was introduced during evaluation. Fig. \ref{fig:win_rate_2d_cn} shows the progression of win-rate during training and evaluation.

The final win-rate achieved during training was \textbf{31.25\%} and during evaluation was \textbf{37.188\%}

\subsection{Experiment 4}
\begin{figure}[h!]
    \centering
    \begin{subfigure}[t]{0.52\textwidth}
        \centering
        \includegraphics[width = \linewidth]{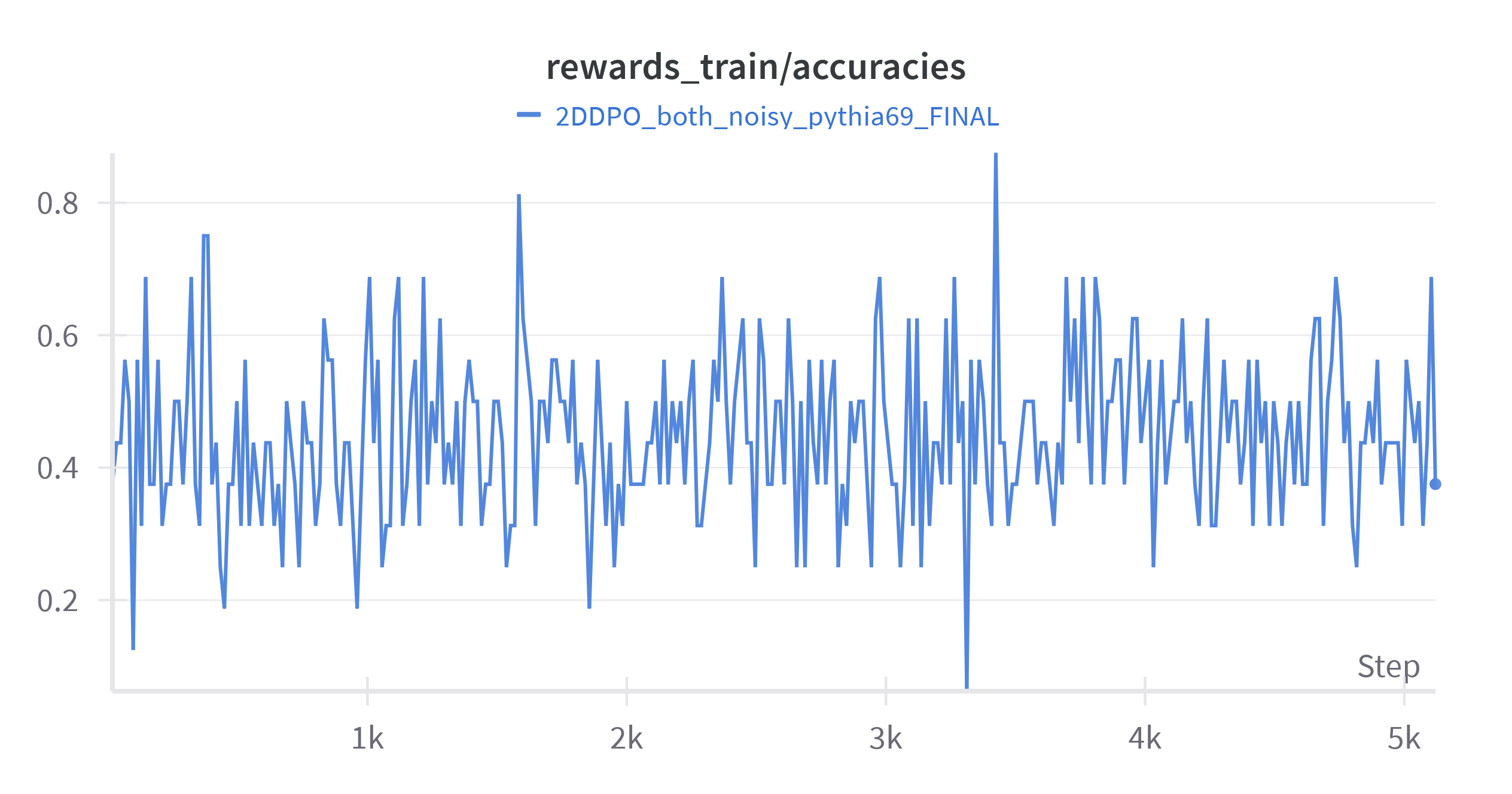}
        \caption{During Training}
    \end{subfigure}%
    ~ 
    \begin{subfigure}[t]{0.52\textwidth}
        \centering
        \includegraphics[width = \linewidth]{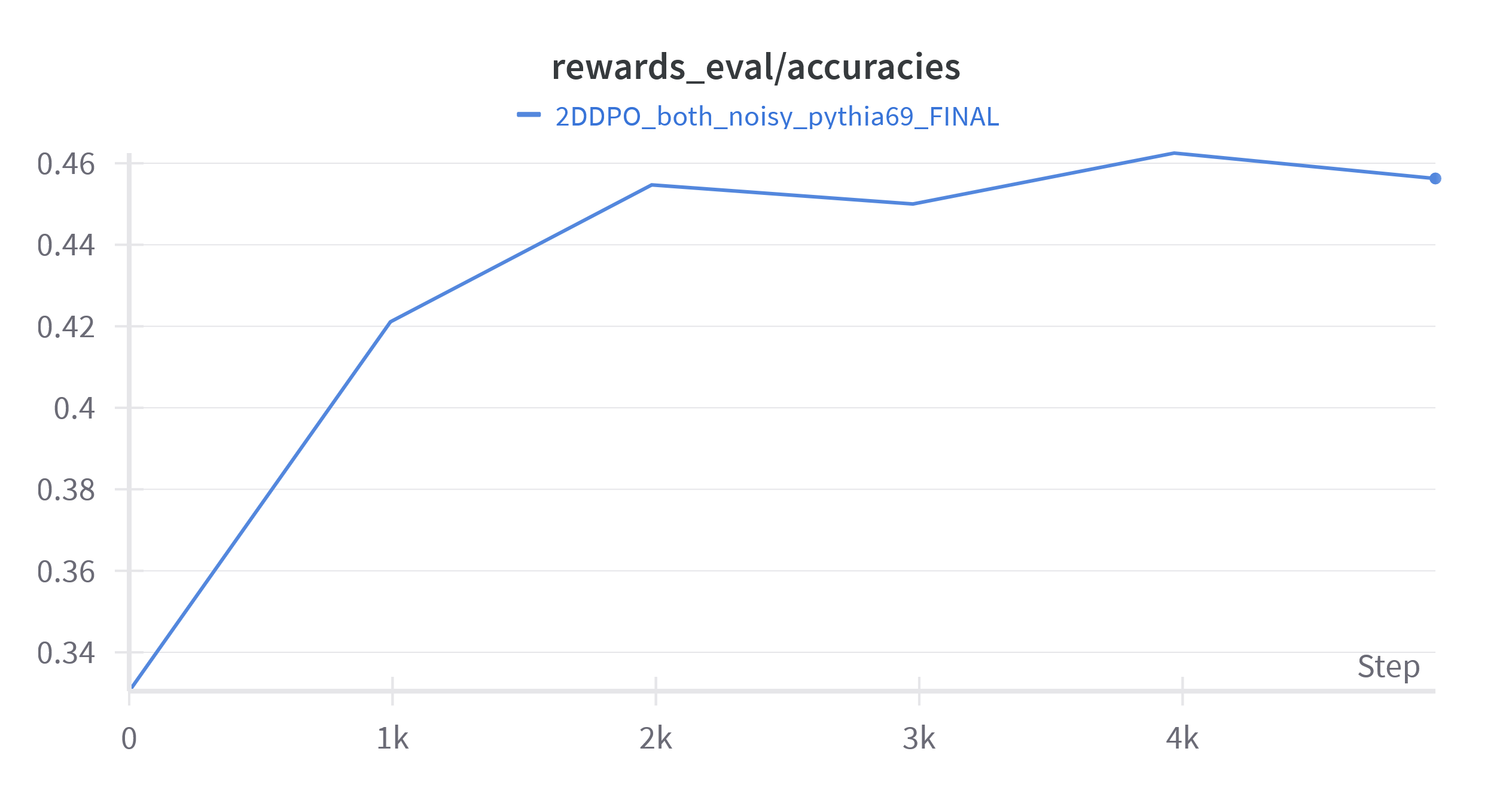}
        \caption{During Evaluation (Noisy)}
    \end{subfigure}%
    \caption{Win rate progression for Robust 2D-DPO}
    \label{fig:win_rate_2d_nn}
\end{figure}

In this experiment, the \textbf{Robust} 2D-DPO algorithm was executed on the \verb|HelpSteer-2D| dataset. Segment level noise had been introduced during the evaluation. Fig. \ref{fig:win_rate_2d_nn} shows the progression of win-rate during training and evaluation.

The final win-rate achieved during training was \textbf{37.188\%} and during evaluation was \textbf{45.625\%}

\subsection{Inferences of Experiments}
After performing the experiments, the final win-rates have been documented in Table \ref{tab:final-res}.

\begin{table}[h]
    \centering
    \begin{tabular}{|c|c|c|}
        \hline
        \textbf{Algorithm} & \textbf{Training Win Rate} & \textbf{Evaluation Win Rate} \\
        \hline
        Vanilla DPO & 58.333\% & 58.854\% \\
        \hline
        Vanilla 2D-DPO & 31.25\% & 43.281\% \\
        \hline
        Vanilla 2D-DPO under noise & 31.25\% & 37.188\% \\
        \hline
        Robust 2D-DPO under noise & 37.188\% & 45.625\% \\
        \hline
    \end{tabular}
    \caption{Consolidated results of Experiments}
    \label{tab:final-res}
\end{table}

As per the results in the Table \ref{tab:final-res}, we observed that under no noise conditions, Vanilla 2D-DPO (\cite{2ddpo}) performed to a satisfactory level (as per the win rate). However, when segment level label noise was introduced, the Vanilla DPO algorithm's win rate experienced a shart decline from 43.281\% to 37.188\%. \\

The above case indicated that Vanilla 2D-DPO is sensitive to noise in the segment scores and is not robust to such perturbations. The results also showed that \textbf{Robust} 2D-DPO, as proposed in this work (Algorithm \ref{alg:1}) was not affected by noise in the system. With a win rate of \textbf{45.625\%}, Robust 2D-DPO was able to handle segment-level perturbations and provide satisfactory performance.

\section{Conclusion and Future Work}
In this work, we proposed a Robust version of 2-dimensional Direct Preference Optimisation which handles noise at the \textbf{segment level}. The noise model used in this work would add a small perturbation to the segment scores during training, where this perturbation was sampled from a uniform distribution. We were able to theoretically model the noisy framework and provide empirical results on the performance of the model under these noisy conditions. Even though this work seems to be a promising step in the direction of robust and granular preference optimisation algorithms, there is still a lot of scope for further improvements. The first direction of extension is to model other noise frameworks and make the 2D-DPO algorithm robust to those. \\

One of the more promising directions could be to induce noise-robustness at an \textbf{aspect level}. There could be situations where with some small probability, the aspect scores could be flipped to another discrete level. This would be deterimental as this would affect the segment scores and further the response scores down the line. Modelling this situation is necessary, but comes with its fair share of problems. Since the reward scores are provided by using external annotators, it is not practical to find the prior distribution of these rewards and this can further interfere with the process of modelling a robust algorithm. \\

Another promising noise framework is to have the log-probabilities of the segments remain intact, but the segment scores be flipped. This framework is closer to our framework in intuition, but it is firmly placed in between our noise model and the noise model proposed in Chowdhury et. al. \cite{robdpo}. Since this situation is a practical possibility, it is important to explore robust models to handle this sort of a noise framework. In addition to these noise frameworks, there can be other areas where noise could be introduced and making robust models for these frameworks can be of practical significance. Additionally, it is also possible to have models where mixture of noise types are present and making a robust model to counter all these types at once is a challenge that might be of interest in the near future.

\begin{ack}
We would like to thank Prof. Ganesh Ramakrishnan, who provided us with the opportunity to pursue this project. This project would not have been possible without his guidance and support. We would also like to extend our gratitude to Suraj Racha, our Teaching Assistant, for his constant help and insights throughout the course of the project. We also thank the BharatGen team for their invaluable assistance in providing the required computational power and technical support to run our computationally heavy project code.
\end{ack}

\bibliographystyle{plainnat}
\bibliography{references.bib}
\newpage
\appendix
Appendix \ref{app:a} covers DPO (Direct Preference Optimisation), while Appendix \ref{app:b} presents the Robust DPO formulation, which introduces noise to simulate flipped preferences and incorporates a conservative loss function. Appendix \ref{app:c} provides the Proof of Lemma 1, detailing the theoretical underpinnings supporting the main results. Finally, Appendix \ref{app:d} describes the Noisy 2D-DPO Algorithm, which extends the DPO approach by incorporating two-dimensional noise to enhance robustness and optimize performance in more complex scenarios. Finally, Appendix \ref{app:e} elaborates on the set-up and results of Experiment 5 as mentioned in Table \ref{tab:exps}.

\section{Proof for Direct Preference Optimisation}\label{app:a}
Rafailov et. al. \cite{dpo} proposed DPO where the reward model was written in terms of the policy as shown in Eq. \ref{eq:a.1}

\begin{equation}\label{eq:a.1}
    {r ^ \ast}(s, {a}) \hspace{0.1cm} = \hspace{0.1cm} \beta \log \left( \frac{{\pi ^ \ast}(a | s)}{{\pi ^ \text{SFT}} (a | s)} \right) \hspace{0.1cm} + \hspace{0.1cm} \beta \log {Z ^ \ast}(s)
\end{equation}

Where ${Z^\ast}(s)$ is the \textbf{partition function}. Substituting from Eq. \ref{eq:a.1} to the standard BTL model as proposed by Bradley et. al. \cite{btl}, we get the preference probabilities as seen in Eq. \ref{eq:a.2}
\begin{equation}\label{eq:a.2}
    \mathbb{P} \left( {a_w} \succ {a_l} \hspace{0.1cm} \vert \hspace{0.1cm} s \right) \hspace{0.1cm} = \hspace{0.1cm} \sigma \left[ \beta \log \left( \frac{{\pi ^ \ast}({a_w} | s)}{{\pi ^ \text{SFT}} ({a_w} | s)} \right) \hspace{0.1cm} - \hspace{0.1cm} \beta \log \left( \frac{{\pi ^ \ast}({a_l} | s)}{{\pi ^ \text{SFT}} ({a_l} | s)} \right) \right]
\end{equation}

We can define the margin of difference between the reward scores for preferred and rejected responses as shown in Eq. \ref{eq:a.3}.
\begin{equation}\label{eq:a.3}
    {h_\theta} \left( s, {a_w}, {a_l} \right) \hspace{0.1cm} = \hspace{0.1cm} {r_\theta} (s, {a_w}) - {r_\theta}(s, {a_l})
\end{equation}

Substituting this back in Eq. \ref{eq:a.2}, we get - 
\begin{equation}\label{eq:a.4}
    \mathbb{P} \left( {a_w} \succ {a_l} \hspace{0.1cm} \vert \hspace{0.1cm} s \right) \hspace{0.1cm} = \hspace{0.1cm} \sigma \left[ \beta {h_\theta} \left( s, {a_w}, {a_l} \right) \right]
\end{equation}
This will enable us to formulate the loss that we wish to minimize - 
\begin{equation}\label{eq:a.5}
    \mathcal{L} \left( \theta \hspace{0.05cm} ; \hspace{0.05cm} s, {a_w}, {a_l} \right) \hspace{0.1cm} = \hspace{0.1cm} - \log \sigma \left( \beta {h_\theta} \left( s, {a_w}, {a_l} \right) \right)
\end{equation}

\section{Robust DPO formulation}\label{app:b}
Given corrupted dataset $\Tilde{\mathcal{D}}$, one can use Eq. \ref{eq:3.3} to get the expression in Eq. \ref{eq:b.1}.
\begin{equation}\label{eq:b.1}
    {\mathcal{L}_\epsilon} \left( \theta \hspace{0.1cm} ; \hspace{0.1cm} s, {\Tilde{a}_w}, {\Tilde{a}_l}\right) \hspace{0.1cm} = \hspace{0.1cm} - \log {\mathbb{P}_{\theta, \epsilon}} \left[ {\Tilde{a}_w} \succ {\Tilde{a}_l} \hspace{0.1cm} \vert \hspace{0.1cm} s \right]
\end{equation}

We know that the preferences are flipped with a small probability of $\epsilon$. Thus, we can write the predicted probabilities under noisy preferences as shown in Eq. \ref{eq:b.2}.
\begin{equation}\label{eq:b.2}
    \begin{split}
        {\mathbb{P}_{\theta, \epsilon}} \left[ {\Tilde{a}_w} \succ {\Tilde{a}_l} \hspace{0.1cm} \vert \hspace{0.1cm} s \right] \hspace{0.1cm} & = \hspace{0.1cm} \left( 1 - \epsilon \right) {\mathbb{P}_\theta} \left[ {\Tilde{a}_w} \succ {\Tilde{a}_l} \hspace{0.1cm} \vert \hspace{0.1cm} s \right] \hspace{0.1cm} + \hspace{0.1cm} \epsilon {\mathbb{P}_\theta} \left[ {\Tilde{a}_l} \succ {\Tilde{a}_w} \hspace{0.1cm} \vert \hspace{0.1cm} s \right] \\
        & = \hspace{0.1cm} \left( 1 - \epsilon \right) \sigma \left( \beta {h_\theta} \left( s, {\Tilde{a}_w}, {\Tilde{a}_l} \right) \right) \hspace{0.1cm} + \hspace{0.1cm} \epsilon \sigma \left( \beta {h_\theta} \left( s, {\Tilde{a}_l}, {\Tilde{a}_w} \right) \right)
    \end{split}
\end{equation}

We know that the logarithmic function is concave, we can write the following expression - 
\begin{equation}\label{eq:b.3}
    \begin{split}
        \log {\mathbb{P}_{\theta, \epsilon}} \left[ {\Tilde{a}_w} \succ {\Tilde{a}_l} \hspace{0.1cm} \vert \hspace{0.1cm} s \right] \hspace{0.1cm} & = \hspace{0.1cm} \log \left\{ \left( 1 - \epsilon \right) \sigma \left( \beta {h_\theta} \left( s, {\Tilde{a}_w}, {\Tilde{a}_l} \right) \right) \hspace{0.1cm} + \hspace{0.1cm} \epsilon \sigma \left( \beta {h_\theta} \left( s, {\Tilde{a}_l}, {\Tilde{a}_w} \right) \right) \right\} \\
        \hspace{0.1cm} & \geq \hspace{0.1cm} \left( 1 - \epsilon \right) \hspace{0.05cm} \log \sigma \left( \beta {h_\theta} \left( s, {\Tilde{a}_w}, {\Tilde{a}_l} \right) \right) \hspace{0.1cm} + \hspace{0.1cm} \epsilon \hspace{0.05cm} \log \sigma \left( \beta {h_\theta} \left( s, {\Tilde{a}_l}, {\Tilde{a}_w} \right) \right)
    \end{split}
\end{equation}

The result from Eq. \ref{eq:b.3} can be used to upper bound the loss since the loss is nothing but the negative of the LHS of Eq. \ref{eq:b.3}. In other words, we can have the noisy loss as - 
\begin{equation}\label{eq:b.4}
    \begin{split}
        {\Tilde{\mathcal{L}}_\epsilon} \left( \theta \hspace{0.1cm} ; \hspace{0.1cm} s, {\Tilde{a}_w}, {\Tilde{a}_l}\right) \hspace{0.1cm} & = \hspace{0.1cm} \left( 1 - \epsilon \right) \hspace{0.05cm} \log \sigma \left( \beta {h_\theta} \left( s, {\Tilde{a}_w}, {\Tilde{a}_l} \right) \right) \hspace{0.1cm} + \hspace{0.1cm} \epsilon \hspace{0.05cm} \log \sigma \left( \beta {h_\theta} \left( s, {\Tilde{a}_l}, {\Tilde{a}_w} \right) \right) \\
        & = \hspace{0.1cm} \left( 1 - \epsilon \right) {\mathcal{L}} \left( \theta \hspace{0.1cm} ; \hspace{0.1cm} s, {\Tilde{a}_w}, {\Tilde{a}_l}\right) + \epsilon {\mathcal{L}} \left( \theta \hspace{0.1cm} ; \hspace{0.1cm} s, {\Tilde{a}_l}, {\Tilde{a}_w}\right)
    \end{split}
\end{equation}

The loss in Eq. \ref{eq:b.4} is called the \textbf{Conservative DPO} loss. However, there is a problem with this formulation. We have - 
\begin{align*}
    \mathbb{E} \left[ {\Tilde{\mathcal{L}}_\epsilon} \left( \theta \hspace{0.1cm} ; \hspace{0.1cm} s, {\Tilde{a}_w}, {\Tilde{a}_l}\right) \right] \hspace{0.1cm} \ne \hspace{0.1cm} \mathbb{E} \left[ \mathcal{L}\left( \theta \hspace{0.1cm} ; \hspace{0.1cm} s, {{a}_w}, {{a}_l}\right) \right]  \\
    \mathbb{E} \left[ {\mathcal{L}_\epsilon} \left( \theta \hspace{0.1cm} ; \hspace{0.1cm} s, {\Tilde{a}_w}, {\Tilde{a}_l}\right) \right] \hspace{0.1cm} \ne \hspace{0.1cm} \mathbb{E} \left[ \mathcal{L}\left( \theta \hspace{0.1cm} ; \hspace{0.1cm} s, {{a}_w}, {{a}_l}\right) \right]  \\
    \mathrm{logit} \left( {\mathbb{P}_{\theta, \epsilon}} \left[ {a_w} \succ {a_l} \hspace{0.1cm} \vert \hspace{0.1cm} s \right] \right) \hspace{0.1cm} \ne \hspace{0.1cm} \mathrm{logit} \left( {\mathbb{P}_{\theta}} \left[ {a_w} \succ {a_l} \hspace{0.1cm} \vert \hspace{0.1cm} s \right] \right)
\end{align*}

The above equations point to the fact that the loss formulation is biased and there is a need to find an unbiased version of this loss formulation. Chowdhury et. al. \cite{robdpo} further propose an unbiased preference probability expression as shown in Eq. \ref{eq:b.5}.
\begin{equation}\label{eq:b.5}
    {\hat{\mathbb{P}}_{\theta, \epsilon}} \left[ {a_w} \succ {a_l} \hspace{0.1cm} \vert \hspace{0.1cm} s \right] \hspace{0.1cm} = \hspace{0.1cm} \frac{\sigma {{\left( \beta {h_\theta} \left( s, {a_w}, {a_l} \right) \right)} ^{(1 - \epsilon)}}}{\sigma {{\left( \beta {h_\theta} \left( s, {a_l}, {a_w} \right) \right)} ^{\epsilon}}}
\end{equation}
Using the definition of the Logit function, we can see that - 
\begin{equation}\label{eq:b.6}
    \mathrm{logit} \left( {\hat{\mathbb{P}}_{\theta, \epsilon}} \left[ {a_w} \succ {a_l} \hspace{0.1cm} \vert \hspace{0.1cm} s \right] \right) \hspace{0.1cm} = \hspace{0.1cm} \mathrm{logit} \left( {\mathbb{P}_{\theta}} \left[ {a_w} \succ {a_l} \hspace{0.1cm} \vert \hspace{0.1cm} s \right] \right)
\end{equation}

\subsection{Sanity Check for Eq. \ref{eq:b.6}}
We know from Eq. \ref{eq:3.3} - 
\begin{equation}
    \label{eq:b.7}
    \mathbb{P}\left( {a_w} \succ {a_l} \hspace{0.1cm} \vert \hspace{0.1cm} s \right) \hspace{0.1cm} = \hspace{0.1cm} \sigma \left( \beta {h_\theta} (s, {a_w}, {a_l}) \right)
\end{equation}

Let us consider the log-odds of the sigmoid function.
\begin{equation}
    \label{eq:b.8}
    \begin{split}
        \mathrm{logit}\left(\sigma (x) \right) \hspace{0.1cm} & = \hspace{0.1cm} \log \left[ \frac{\sigma(x)}{1 - \sigma(x)} \right] \hspace{0.1cm} = \hspace{0.1cm} \log \left[ \frac{\sigma(x)}{\sigma(-x)}\right] \\
        \hspace{0.1cm} & = \hspace{0.1cm} \log \left[ \frac{\frac{\exp{(x)}}{1 + \exp{(x)}}}{\frac{1}{1 + \exp{(x)}}}\right] \hspace{0.1cm} = \hspace{0.1cm} \log \left[ \frac{\exp{(x)}}{1}\right] \\
        \hspace{0.1cm} & = \hspace{0.1cm} x
    \end{split}
\end{equation}

Using the result in Eq. \ref{eq:b.8} in Eq. \ref{eq:b.7}, we get - 
\begin{equation}
    \label{eq:b.9}
    \begin{split}
        \mathrm{logit} \left[ \mathbb{P}\left( {a_w} \succ {a_l} \hspace{0.1cm} \vert \hspace{0.1cm} s \right) \right] \hspace{0.1cm} & = \hspace{0.1cm} \mathrm{logit} \left[ \sigma \left( \beta {h_\theta} (s, {a_w}, {a_l}) \right) \right] \\
        \hspace{0.1cm} & = \hspace{0.1cm} \beta {h_\theta} (s, {a_w}, {a_l})
    \end{split}
\end{equation}

Let us assume that 
\begin{equation}
    \label{eq:b.10}
    \beta {h_\theta} (s, {a_w}, {a_l}) = z \hspace{0.1cm} \mathrm{(say)}
\end{equation}

Then, Eq. \ref{eq:b.5} becomes - 
\begin{equation}
    \label{eq:b.11}
    \begin{split}
        {\hat{\mathbb{P}}_{\theta, \epsilon}} \left[ {a_w} \succ {a_l} \hspace{0.1cm} \vert \hspace{0.1cm} s \right] \hspace{0.1cm} & = \hspace{0.1cm} \frac{\sigma {{\left( \beta {h_\theta} \left( s, {a_w}, {a_l} \right) \right)} ^{(1 - \epsilon)}}}{\sigma {{\left( \beta {h_\theta} \left( s, {a_l}, {a_w} \right) \right)} ^{\epsilon}}} \\
        \hspace{0.1cm} & = \hspace{0.1cm} \frac{\sigma {{\left( \beta {h_\theta} \left( s, {a_w}, {a_l} \right) \right)} ^{(1 - \epsilon)}}}{\sigma {{\left( -\beta {h_\theta} \left( s, {a_w}, {a_l} \right) \right)} ^{\epsilon}}} \\
         \hspace{0.1cm} & = \hspace{0.1cm} \frac{{\sigma (z)}^{(1 - \epsilon)}}{{\sigma (-z)}^{\epsilon}} \hspace{0.1cm} = \hspace{0.1cm} \frac{{\sigma (z)}^{(1 - \epsilon)}}{\left[{1 - \sigma (z)}\right]^{\epsilon}} \\
         \hspace{0.1cm} & = \hspace{0.1cm} A \hspace{0.1cm} \mathrm{(say)}
    \end{split}
\end{equation}

After defining $A$ in Eq. \ref{eq:b.11}, we can write - 
\begin{equation}
    \label{eq:b.12}
    \begin{split}
        \frac{A}{1 - A} \hspace{0.1cm} & = \hspace{0.1cm} \frac{\frac{{\sigma (z)}^{(1 - \epsilon)}}{\left[{1 - \sigma (z)}\right]^{\epsilon}}}{ 1 - \frac{{\sigma (z)}^{(1 - \epsilon)}}{\left[{1 - \sigma (z)}\right]^{\epsilon}}}
    \end{split}
\end{equation}

This shall motivate the formulation of an unbiased loss function that has been written in Eq. \ref{eq:3.7}.

\section{Proof of Lemma \ref{lem:4.1}}
\label{app:c}
Consider a value $x \in \mathbb{R}$ and the standard sigmoid function - 
\[
    \sigma(x) \hspace{0.1cm} = \hspace{0.1cm} \frac{1}{1 + \exp{(-x)}} \hspace{0.1cm} = \hspace{0.1cm} \frac{\exp{(x)}}{1 + \exp{(x)}}
\]

\textbf{\underline{To prove necessity}} \\
We need to show that - 
\begin{equation}
    \label{eq:c.1}
    \log (\sigma(x)) = \log(\sigma(-x)) \hspace{0.3cm} \implies \hspace{0.3cm} x = 0
\end{equation}

Expanding the sigmoid in Eq. \ref{eq:c.1}, we get - 
\begin{equation}
    \label{eq:c.2}
    \begin{split}
        \log \left[\frac{\exp{(x)}}{1 + \exp{(x)}} \right] \hspace{0.1cm} & = \hspace{0.1cm} \log \left[\frac{\exp{(-x)}}{1 + \exp{(-x)}} \right] \\
        \log \left[\frac{\exp{(x)}}{1 + \exp{(x)}} \right] \hspace{0.1cm} & = \hspace{0.1cm} \log \left[\frac{1}{1 + \exp{(x)}} \right]
    \end{split}
\end{equation}

Applying the monotonicity property of the $\log$, we get - 
\begin{equation}
    \label{eq:c.3}
    \begin{split}
        \frac{\exp{(x)}}{1 + \exp{(x)}} \hspace{0.1cm} & = \hspace{0.1cm} \frac{1}{1 + \exp{(x)}} \\
        \exp(x) \hspace{0.1cm} & = \hspace{0.1cm} 1
    \end{split}
\end{equation}
Thus, the only possible value $x$ can take is $x = 0$ \\

\textbf{\underline{To prove sufficiency}} \\
We need to show that - 
\begin{equation}
    \label{eq:c.4}
    x = 0 \hspace{0.3cm} \implies \hspace{0.3cm} \log (\sigma(x)) = \log(\sigma(-x))
\end{equation}

Substituting $x = 0$ in the sigmpoid function, we get - 
\[
    \sigma(x = 0) \hspace{0.1cm} = \hspace{0.1cm} \frac{\exp{(0)}}{1 + \exp{(0)}} \hspace{0.1cm} = \hspace{0.1cm} \frac{1}{2}
\]
Similarly, we have - 
\[
    \sigma (x = -0) \hspace{0.1cm} = \hspace{0.1cm} \frac{1}{1 + \exp{(0)}} \hspace{0.1cm} = \hspace{0.1cm} \frac{1}{2}
\]

Thus, we can say that if $x = 0$, then we get $\log (\sigma(x)) = \log(\sigma(-x))$
\section{Noisy 2D-DPO Algorithm}
\label{app:d}
\begin{algorithm}[H]
    \caption{Noisy 2D-DPO Optimisation via Mini Batch GD}\label{alg:1}
    \begin{algorithmic}[1]
        \State \textbf{Input : } Parameters $\theta$, Policy model $\pi_{\theta}$, Reference policy $\pi_{\mathrm{ref}}$, Temperature $\beta > 0$, Learning rate $\eta > 0$, Mini-batch size $B$, Number of training iterations $T$
        \State $\theta \gets \theta_0$
        \For{ iteration from $1$ to $T$ }
            \State Initialize: Mini batch $\mathcal{B} = \left\{ {\tau_{w}^{(i)}}, {\tau_{l}^{(i)}}, {r_{w}^{(i)}}, {r_{l}^{(i)}} \right\}_{i=1}^B$, Gradient Accumulator ${g_{\mathrm{batch}}} \gets 0$
            \For{ $i$ from $1$ to $B$ }
                \State Sample Noise $\delta \sim U(0,1)$, Initialize Sample Loss ${L_{\mathrm{sample}}} \gets 0$
                \For{ $k$ from $0$ to ${n^{(i)}} - 1$}
                    \State $l_{\left(w,k\right)} \gets \beta {\sum_{j = {n_k}}^{{n_k} + {l_k}}} \left[ \log {\pi_\theta} \left( {a_j} \vert {s_j} \right) \hspace{0.1cm} - \hspace{0.1cm} \log {\pi_{\mathrm{ref}}} \left( {a_j} \vert {s_j} \right) \right]$
                    \State $l_{\left(l,k\right)} \gets \beta {\sum_{j = {n_k}}^{{n_k} + {l_k}}} \left[ \log {\pi_\theta} \left( {a_j} \vert {s_j} \right) \hspace{0.1cm} - \hspace{0.1cm} \log {\pi_{\mathrm{ref}}} \left( {a_j} \vert {s_j} \right) \right]$
                    \State ${X_k} \gets {r_{w}^{(i)}}l_{\left(w,k\right)} - {r_{l}^{(i)}}l_{\left(l,k\right)}$
                    \State ${Y_k} \gets l_{\left(w,k\right)} + l_{\left(l,k\right)}$
                    \State ${L_{\mathrm{sample}}} \gets {L_{\mathrm{sample}}} + \log \left( \sigma \left\{ {X_k} - \delta {Y_k}\right\} \right)$
                \EndFor
                \State ${g_{\mathrm{batch}}} \gets {g_{\mathrm{batch}}} + {\nabla_\theta}{L_{\mathrm{sample}}}$
            \EndFor
            \State ${g_{\mathrm{avg}}} \gets {g_{\mathrm{batch}}} / B$
            \State $\theta \gets \theta - \eta {g_{\mathrm{avg}}}$
        \EndFor
        \State \textbf{Return} $\theta$
    \end{algorithmic}
\end{algorithm}

\section{Experiment 5 results}
\label{app:e}
The first stage of experimentation was to implement Vanilla DPO algorithm using the code provided by Rafailov et al. \cite{dpo} \footnote{\href{https://github.com/eric-mitchell/direct-preference-optimisation}{GitHub-code}}. This would help us check the sanity of the code and implementation and ensure smooth transition to implementing 2D-DPO. Table \ref{tab:dpo-setup} contains the experiment details for running Vanilla DPO.

\begin{table}[h]
    \centering
    \begin{tabular}{|m{2cm}|m{6cm}|}
        \hline
        \textbf{Model} & Pythia - 2.8B \\
        \hline
        \textbf{Dataset} & Anthropic-HH (161k train split, 8552 test split) Bai et al. \cite{anthropichh} \\
        \hline
        \multirow{2}{*}{\textbf{Time taken}} & SFT - 220 mins \\
        \cline{2-2}
        & DPO - 280 mins \\
        \hline
    \end{tabular}
    \caption{Experimental Details for Vanilla DPO}
    \label{tab:dpo-setup}
\end{table}

\textbf{Anthropic-HH} is a human preference dataset about helpfulness and harmlessness from . As per the data format, each line of the JSONL file contains a pair of responses: one ``chosen'' and one ``rejected''. For \textbf{helpfulness}, the data is grouped into train/test splits across three groups: 
\begin{itemize}
    \item Context-distilled 52B language models.
    \item Rejection sampled data points (mostly with best-of-16 sampling) against an early preference model.
    \item Dataset sampled during the author's iterated ``online" process.
\end{itemize}

For \textbf{harmlessness}, the data is collected in a similar fashion, but only for the base models. \\

\begin{figure*}[h!]
    \centering
    \begin{subfigure}[t]{0.5\textwidth}
        \centering
        \includegraphics[width = 1.1\linewidth]{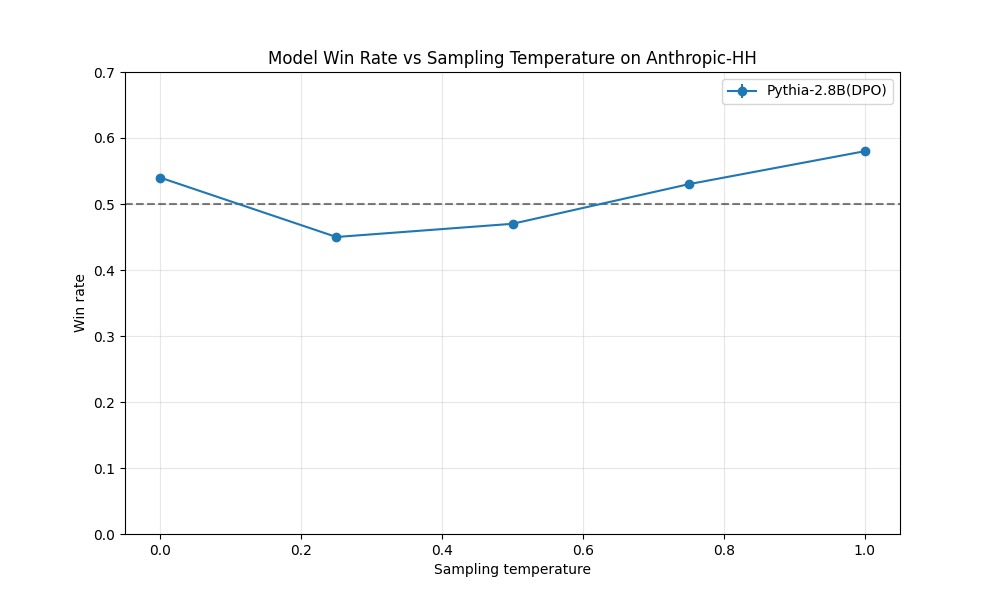}
        \caption{For 100 examples}
    \end{subfigure}%
    ~ 
    \begin{subfigure}[t]{0.5\textwidth}
        \centering
        \includegraphics[width = 1.1\linewidth]{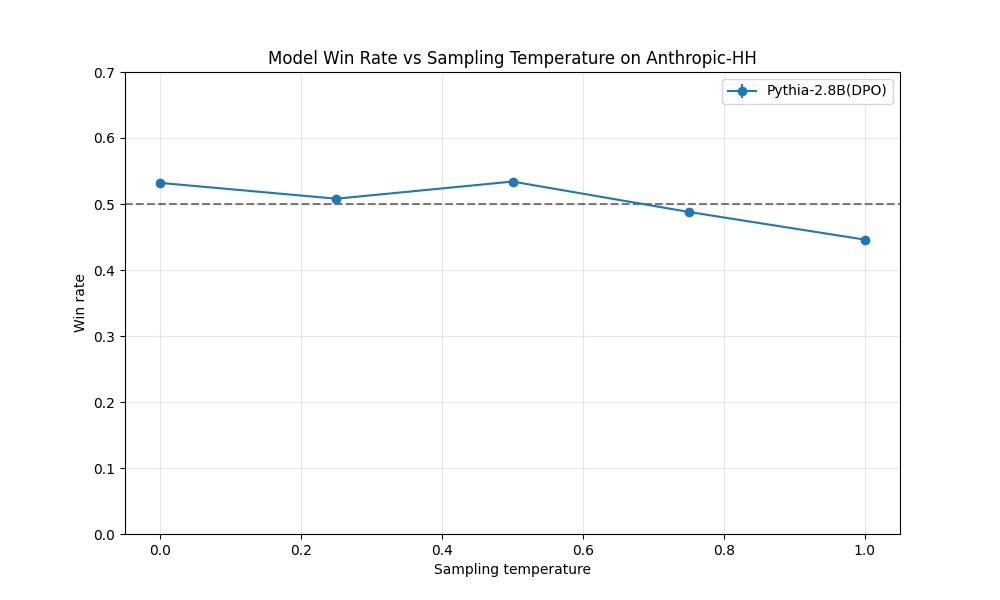}
        \caption{For 500 examples}
    \end{subfigure}
    \caption{Win rate vs. Sampling Temperature}
    \label{fig:dpo}
\end{figure*}

The preference optimized model was first evaluated against 100 prompt and then against 500 prompts. The analysis involved observing the win rate against the \textbf{sampling temperature}, where the win rate is calculated as the percetange of test prompts for which the model is able to correctly distinguish between preferred and rejected responses. Fig. \ref{fig:dpo} shows the change in win rates as the sampling temperature is varied. 

We observe that the findings are in-line with the work done by Rafailov et al. \cite{dpo}. 

\end{document}